\documentclass[runningheads]{llncs}
\usepackage{amsfonts}
\usepackage{amsmath}
\usepackage[T1]{fontenc}
\usepackage{graphicx,verbatim}
\usepackage{hyperref}
\usepackage{color}

\usepackage{subcaption}

\usepackage{multirow}
\usepackage{tabularx}
\usepackage{cite}
\usepackage{svg}
\usepackage[normalem]{ulem}
\useunder{\uline}{\ul}{}
\newcolumntype{C}{>{\centering\arraybackslash}X}

\begin{document}
\title{Step-Attention Refinement of DINOv3 Features for Efficient Anterior Eye Segmentation}
\titlerunning{Step-Attention Refinement of DINOv3 Features for AES Segmentation}
%
\author{
Philippe Baumstimler \inst{1} \and
Jean-Mathieu Gagnon \inst{2} \and
Sébastien Gagné \inst{2,3} \and
Mathieu Duchesneau \inst{2} \and
Clément Playout \inst{1} \and
Lama Séoud \inst{1}
}
\authorrunning{P. Baumstimler et al.}
%
\institute{Polytechnique Montréal, Montréal QC, Canada \and
LightX Innovations, Montréal QC, Canada \and
Institut de l'Oeil des Laurentides, Boisbriand QC, Canada}


  
\maketitle              
\begin{abstract}

Anterior eye segment (AES) segmentation is a key component of both ocular biometrics and emerging clinical image analysis applications. However, heterogeneous acquisition conditions and limited annotations in medical settings hinder the robustness and generalization of existing methods. Foundation models (FMs) such as DINOv3 offer strong transfer capabilities, but efficiently adapting their representations to dense prediction tasks remains challenging. In this study, we investigate robust AES segmentation in clinical settings, and propose a lightweight architecture built upon a distilled DINOv3 ViT-Small backbone. We introduce a step-attention feature refinement module that progressively adapts multi-level transformer representations before convolutional decoding, enabling efficient exploitation of pretrained features with few parameters. We evaluate the proposed approach on a private dataset of 333 clinically acquired AES images spanning eight ophthalmic acquisition protocols and annotated for seven anatomical classes. Compared with convolutional and transformer-based baselines, including DINOv3-based methods, our approach achieves the best overall performance, reaching 85.55\% mIoU when fully fine-tuned. It also demonstrates the strongest robustness to domain shift across four unseen public AES segmentation datasets. These results establish a strong baseline for robust AES segmentation in clinical settings and highlight the importance of decoder design for effectively adapting FMs representations to medical segmentation tasks.


\end{abstract}

\section{Introduction}

Semantic segmentation of anterior eye segment (AES) photographs is a key component in applications such as gaze estimation in human–machine interaction \cite{chaudhary2019ritnet} and iris or sclera recognition in ocular biometrics \cite{mobius, ssbc2025}. These efforts have led to several public datasets \cite{mobius, sbvpi, casia_irisv4}, mainly targeting sclera, iris, and pupil segmentation under controlled acquisition settings.

More recently, AES segmentation has emerged as an important component of medical image analysis pipelines by providing structural priors for clinically meaningful measurements and region-specific biomarkers. For instance, Nahass et al. \cite{Nahass2025} introduced a dedicated segmentation dataset for periorbital distance prediction in oculoplastic and craniofacial surgery. Segmentation of the palpebral conjunctiva has proven to improve haemoglobin estimation and anemia screening \cite{eda, CAMPOREALE2025111026}, while scleral and conjunctival vessel segmentation has enabled automatic conjunctival hyperemia grading \cite{Wong2025}. More broadly, anatomical segmentation can support the development of interpretable models for ocular disease analysis by explicitly localizing clinically meaningful regions \cite{Visionome}.

However, robust AES segmentation remains challenging. Although several studies have reported promising cross-dataset transfer capabilities \cite{mobius, ssbc2025}, degradation under domain shift is still observed. Moreover, current domain-gap evaluations are mostly restricted to biometric datasets and limited to sclera, iris and pupil segmentation. In contrast, clinical AES imaging exhibits stronger variability in acquisition conditions,  including  occlusions, illumination changes, viewpoint, focus and imaging devices, and requires segmentation of less-studied regions such as the palpebral conjunctiva and caruncle for which annotations remain scarce \cite{eda, slid, Nahass2025}. As a result, developing segmentation models that generalize across acquisition settings and anatomical targets remains an open challenge.

Foundation models (FMs) have recently demonstrated strong representation learning capabilities. In the vision-only paradigm, the DINO family \cite{caron2021dino}, and in particular its latest iteration DINOv3 \cite{simeoni2025dinov3}, has exhibited remarkable zero-shot performance from classification to segmentation tasks. Nevertheless, adapting ViT representations to dense prediction remains non-trivial. While SegDINO \cite{yang2025segdino} relies on a lightweight MLP decoder, such a design may lack the discriminative capacity required for complex multi-class segmentation tasks. Conversely, architectures specifically designed for dense prediction from ViT features, such as DPT \cite{Ranftl2021dpt}, generally achieve stronger performance but at the cost of increased computational complexity.

In this work, we address AES semantic segmentation in clinical environments. Inspired by SegDINO \cite{yang2025segdino} and DPT \cite{Ranftl2021dpt}, we propose a lightweight step-attention feature refinement module built upon a distilled DINOv3 ViT-Small backbone \cite{simeoni2025dinov3}. Given that AES segmentation primarily involves large, low-frequency structures, we hypothesize that accurate segmentation relies more on semantically rich coarse
representations than complex high-resolution  decoding. Our design results in robust AES segmentation performance with few parameters by progressively refining DINOv3 representations through attention-based feature adaptation prior to convolutional upsampling. Our contributions are listed as follow: 1) We constructed a private clinically acquired AES segmentation dataset comprising more than 300 manually annotated images collected from eight ophthalmic acquisition protocols, with seven clinically and biometrically relevant anatomical structures and acquisition artifacts. 2) We introduce a lightweight DINOv3-based architecture specifically designed for coarse anatomical segmentation and featuring a custom step-attention feature refinement module that progressively adapts transformer representations before convolutional decoding. 3) We conduct a comprehensive evaluation against state-of-the-art baselines on both in-domain and cross-domain settings, including four unseen AES segmentation public datasets, demonstrating the robustness and generalization capabilities of the proposed approach under domain shifts. 

\section{Methodology}

\begin{figure}[t]
    \centering
    \includegraphics[width=0.8\linewidth]{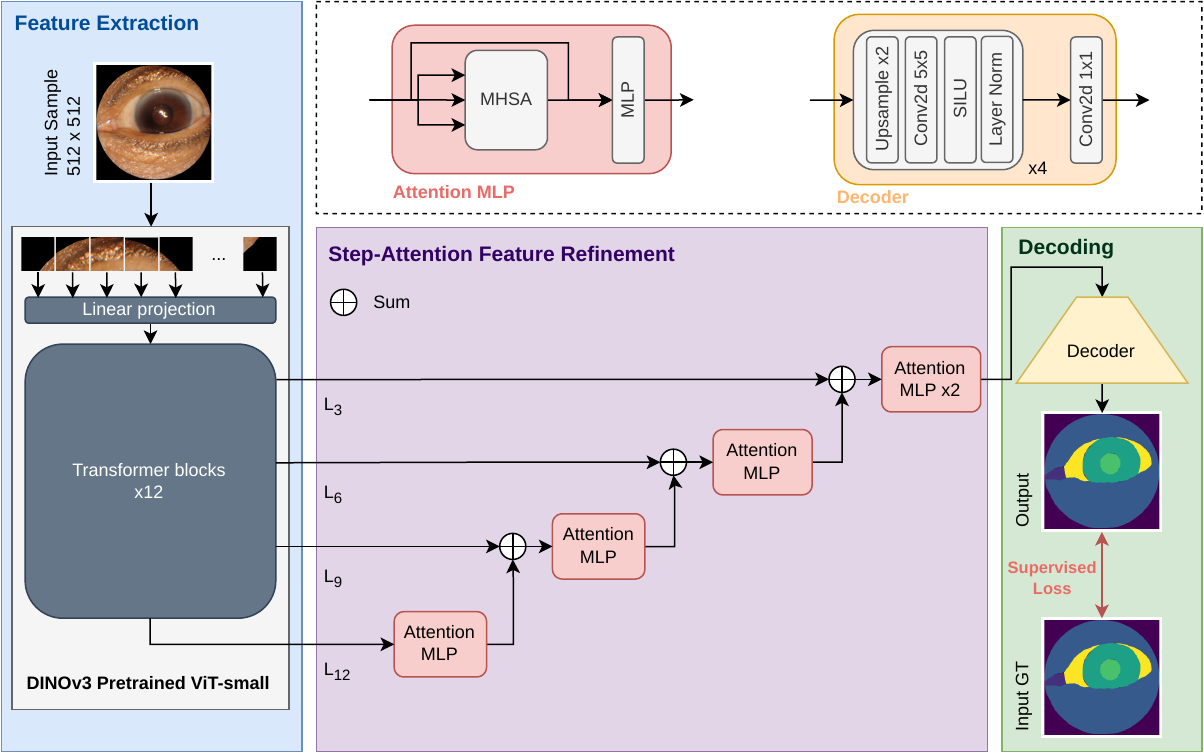}
    \caption{Overview of the proposed model architecture based on DINOv3 pre-trained ViT for coarse segmentation tasks.}
    \label{fig:arch}
\end{figure}

In this section, we present the proposed architecture, shown in Figure \ref{fig:arch}. The model follows a standard encoder–decoder design based on a DINOv3 pre-trained Vision Transformer. Features from multiple layers are progressively fused and refined in the low-resolution latent space through attention-based adaptation modules before convolutional upsampling. This design enables expressive feature refinement for coarse segmentation tasks.

\subsection{Feature Extraction}

As backbone, we employ a ViT pre-trained using the DINOv3 self-supervised learning framework on the LVD-1689M dataset \cite{simeoni2025dinov3}. Feature extraction follows the protocol proposed in SegDINO \cite{yang2025segdino}.

Let $\mathcal{D}=\{(x_i,y_i)\}_{i\in \mathbb{N}}$ be a dataset of RGB images $x_i\in\mathbb{R}^{H\times W\times 3}$ and their corresponding segmentation masks $y_i\in\mathbb{R}^{H\times W\times C}$, where H and W are respectively the image height and width, while C is the number of segmentation classes. Following the standard ViT formulation, inputs are first divided into $N=\frac{HW}{p^2}$ non-overlapping square patches of size $p$. Each of them is then linearly projected into a d-dimension token embedding, forming the initial sequence $z_0\in\mathbb{R}^{N\times d}$. As proposed in DINOv3 \cite{simeoni2025dinov3}, a class token and four additional register tokens are concatenated to the input sequence, yielding $z_0'\in\mathbb{R}^{(N+5)\times d}$. Unlike conventional ViTs relying on traditional learnable positional embeddings, the DINOv3 ViT adopts the more recent rotary position embedding (RoPE) \cite{rope}, which encodes relative spatial information directly within the self-attention computation by applying a rotational transformations to the contextual representation. Token embeddings are then sequentially processed by a stack of $L$ transformer blocks. Let $f_l$ be the l-th transformer block, each intermediate representation $z_l$ can be defined as:

\begin{equation}
    \forall l\in\{1, \dots, L\}, 
    z'_l = f_l(z'_{l-1})
\end{equation}

Following the hierarchical intuition of UNet \cite{unet}, we extract features from multiple layers to leverage both low-level and high-level patch semantics. For a chosen subset of K layers $\mathcal{L}\subset\{l_1, \dots, l_L\}$, patch tokens are retained while non-patch related tokens are discarded, resulting in the multi-feature set
\begin{equation}
    \mathcal{F} = \{z_k\}_{k=1}^K
\end{equation}
which serves as input to the proposed step-attention feature refinement module. 

\subsection{Step-Attention Feature Refinement}

Given the anatomical characteristics of the eye and the selected region of interests, segmentation primarily involves large, low-frequency structures with limited fine-grained details, except in cases such as eyelash occlusions. We therefore hypothesize that accurate segmentation relies more on semantically rich deep representations than on complex high-resolution decoding. Therefore, instead of adopting the fully convolutional hierarchical decoder proposed in DPT\cite{Ranftl2021dpt}, we introduce a step-attention refinement module that operates directly in the low-resolution feature space.

Inspired by the transformer architecture, each refinement stage consists of a multi-head self-attention (MHSA) layer followed by a multi-layer perceptron (MLP). The MHSA module enables the modeling of long-range spatial dependencies, while the MLP adapts the pre-trained DINOv3 representations to the target segmentation task. To preserve positional information throughout the refinement process, RoPE is applied within each MHSA block, following the same process used in the encoder. Let $g_i:\mathbb{R}^{N\times d}\mapsto \mathbb{R}^{N\times d}$ denote the i-th refinement block. For an input feature patch sequence $z\in\mathbb{R}^{N\times d}$, the refinement block is defined as

\begin{equation}
    g_i(z) = \mathrm{MLP}(z + \mathrm{MHSA}(z)) \\
\end{equation}

To incorporate hierarchical information, representations extracted from different transformer layers are progressively fused in a deep-to-shallow manner. Starting from the deepest feature map, each refinement stage combines the last refined representation with the next shallower feature patch sequence before applying the current block. After all transformer features have been fused, a final refinement block is applied to further adapt the representation to the segmentation task. In total, the refinement module therefore comprises $K+1$ blocks. Let $r_j$ be the refined representation at stage $j$. The overall refinement process can be written as  

\begin{equation}
    \forall j\in\{1, \dots, K+1\}, 
    r_j = 
    \begin{cases}
        g_j(z_K) & \text{if } j=1 \\
        g_j(r_{j-1} + z_{K-j+1}) & \text{if } 1<j<K+1\\
        g_j(r_{j-1}) & \text{if } j=K+1
    \end{cases}
\end{equation}
where $\{z_k\}_{k=1}^{K}$ are the extracted features ordered from shallowest to deepest, with $z_K$ denoting the deepest representation (Figure \ref{fig:arch}). 

This progressive refinement strategy combines the semantic richness of deep DINOv3 representations with the spatial information retained in shallower layers, while maintaining all computations in the low-resolution feature space. Only the last refined representation is kept for final decoding.

\subsection{Convolutional Upsampling}

At this stage, the refined representation remains in the low-resolution feature space. To introduce local spatial dependencies during decoding, we pass the final refined representation to a lightweight four-layer convolutional decoder $D_\theta:\mathbb{R}^{N\times d}\mapsto\mathbb{R}^{H \times W \times C}$, which produces the final segmentation map $\hat{y}$:

\begin{equation}
    D_\theta(r_{K+1}) = \hat{y}\
\end{equation}

The input features are first reshaped into 2D spatial feature maps. Each decoder stage consists of bilinear upsampling followed by a $5\times5$ convolution, SiLU activation, and layer normalization. To reduce computational cost, the embedding dimension is halved after each convolutional block. The final feature map is projected into the label space using a $1\times1$ convolution followed by a softmax activation to produce per-pixel class probabilities.

This design progressively restores spatial resolution while incorporating local spatial correlations, enabling a smooth reconstruction of the segmentation map from the low-resolution refined feature space.

\section{Experiments}

\begin{table}[t]
\centering
\caption{Global performance comparisons of our approach against baselines on in- and out-of-distribution data. Metrics were computed as the macro average across available classes in each datasets over 5 random seeds. Parameter counts refer to trainable parameters only. Best results are shown in \textbf{bold} while second and third bests are \underline{underlined}. -$^\dagger$Frozen encoder weights -$^\star$ Data not seen during training.}
\label{tab:global-perf}
\resizebox{\columnwidth}{!}{%
\begin{tabularx}{1.5\textwidth}{CCC|CC|CC|CC|CC|CC}
\hline
 &
   &
   &
  \multicolumn{2}{c|}{LightX} &
  \multicolumn{2}{c|}{Eyes-defy-anemia$^\star$} &
  \multicolumn{2}{c|}{SLID$^\star$} &
  \multicolumn{2}{c|}{MOBIUS$^\star$} &
  \multicolumn{2}{c}{SBVPI$^\star$} \\ 
\multicolumn{1}{c}{Model} &
  \multicolumn{1}{c}{Backbone} &
  \multicolumn{1}{c|}{Param.} &
  \multicolumn{1}{c}{mIoU $\uparrow$} &
  \multicolumn{1}{c|}{DSC $\uparrow$} &
  \multicolumn{1}{c}{mIoU $\uparrow$} &
  \multicolumn{1}{c|}{DSC $\uparrow$} &
  \multicolumn{1}{c}{mIoU $\uparrow$} &
  \multicolumn{1}{c|}{DSC $\uparrow$} &
  \multicolumn{1}{c}{mIoU $\uparrow$} &
  \multicolumn{1}{c|}{DSC $\uparrow$} &
  \multicolumn{1}{c}{mIoU $\uparrow$} &
  \multicolumn{1}{c}{DSC $\uparrow$} 
\\ \hline \hline
    \multicolumn{1}{c}{UNet}       & 
    \multicolumn{1}{c}{EfficientNet-b0} & 
    6.3M  &  
    77.84   & 
    84.23          & 
    38.07 & 48.70 & 
    85.16 & 
    91.09 & 
    41.57 & 
    52.40 & 
    75.30 & 
    83.01 
\\
    \multicolumn{1}{c}{UNet}       & 
    \multicolumn{1}{c}{EfficientNet-b5} & 
    31.2M & {\ul82.14}    & 
    {\ul87.87}    &
    37.31 &
    48.86 & 
    \textbf{86.09} & 
    \textbf{91.80} & 
    44.49 & 
    54.80  & 
    75.70 &  
    83.18
\\
    \multicolumn{1}{c}{Segformer}  & 
    \multicolumn{1}{c}{mit-b1}          & 
    13.7M  & 
    80.22          & 
    86.34          & 
    54.42 & 
    69.20 & 
    83.60 & 
    90.12 & 
    40.40 & 
    52.23 & 
    75.36 &  
    83.62
\\
    \multicolumn{1}{c}{Segformer}  & 
    \multicolumn{1}{c}{mit-b3}          & 
    44.6M  & 
    {\ul 82.45}          & 
    {\ul 87.93}          & 
    59.09 & 
    72.80 & 
    84.10 & 
    90.46 & 
    41.23 & 
    52.15 & 
    {\ul76.74} &  
    {\ul84.93}
\\
    \multicolumn{1}{c}{DPT} & 
    \multicolumn{1}{c}{ViT-Small}       & 
    52.2M & 
    79.68 & 
    86.27 & 
    {\ul62.40} & 
    {\ul75.15} & 
    {\ul85.50} & 
    {\ul91.41} & 
    {\ul55.82} & 
    {\ul66.83} & 
    74.51 & 
    82.16 
\\ 
    \multicolumn{1}{c}{SegDINO$^\dagger$} & 
    \multicolumn{1}{c}{ViT-Small} & 
    9.5M & 
    70.24 & 
    78.57 & 
    41.22 & 
    54.87 & 
    81.38 & 
    88.90 & 
    {\ul53.45} & 
    {\ul65.83} &  
    71.69 & 
    79.76 
\\ 
\hline

    \multicolumn{1}{c}{Ours$^\dagger$}      & 
    \multicolumn{1}{c}{ViT-Small}       & 
    11.3M  & 
    79.09          & 
    85.74   & 
    {\ul62.53} & 
    {\ul75.88} & 
    84.43 & 
    90.76& 
    52.92 & 
    64.52 & 
    {\ul77.71} & 
    {\ul85.60} 
\\
    \multicolumn{1}{c}{Ours}       & 
    \multicolumn{1}{c}{ViT-Small}       & 
    32.9M & 
    \textbf{85.55} & 
    \textbf{90.42} & 
    \textbf{63.63} & 
    \textbf{77.20} & 
    {\ul85.53} & 
    {\ul91.41} & 
    \textbf{59.61} & 
    \textbf{69.90} & 
    \textbf{79.79} & 
    \textbf{87.51} 
\\
\hline
\end{tabularx}%
}
\end{table}

\begin{table}[t]
\centering
\caption{Per-class mIoU comparisons of our approach against baselines on in-distribution data averaged over 5 random seeds with their standard deviation. Per-class training pixel ratios (excluding background) are reported below each class. Parameter counts refer to trainable parameters only. Best results are shown in \textbf{bold} while second and third bests are \underline{underlined} -$^\dagger$Frozen encoder weights}
\label{tab:per-class}
\resizebox{\columnwidth}{!}{%
\begin{tabularx}{1.5\textwidth}{CCC|CCCCCCCCCCCCCC}
\hline 
  \multicolumn{3}{C|}{} &
  \multicolumn{14}{c}{Results on LightX - mIoU $\uparrow$} 
\\
\multicolumn{1}{C}{Model} &
  \multicolumn{1}{C}{Backbone} &
  \multicolumn{1}{C|}{Param.} &
  \multicolumn{2}{C}{Pupil} &
  \multicolumn{2}{C}{Iris} &
  \multicolumn{2}{C}{Sclera} &
  \multicolumn{2}{C}{Caruncle} &
  \multicolumn{2}{C}{Palp. conj.} &
  \multicolumn{2}{C}{Periocular} &
  \multicolumn{2}{C}{Artifacts}
\\
  \multicolumn{3}{c|}{} &
  \multicolumn{2}{c}{(4.6\%)} &
  \multicolumn{2}{c}{(29.9\%)} &
  \multicolumn{2}{c}{(18.8\%)} &
  \multicolumn{2}{c}{(0.3\%)} &
  \multicolumn{2}{c}{(1.9\%)} &
  \multicolumn{2}{c}{(29.5\%)} &
  \multicolumn{2}{c}{(1.0\%)} 
\\\hline \hline
    \multicolumn{1}{c}{UNet}       & 
    \multicolumn{1}{c}{EfficientNet-b0} & 
    \multicolumn{1}{c|}{6.3M}  &  
    \multicolumn{2}{c}{89.8$\pm$0.5}  & 
    \multicolumn{2}{c}{\underline{91.8$\pm$0.4}}  & 
    \multicolumn{2}{c}{81.8$\pm$0.5}  & 
    \multicolumn{2}{c}{52.2$\pm$2.6}  & 
    \multicolumn{2}{c}{76.0$\pm$2.6}  & 
    \multicolumn{2}{c}{90.0$\pm$0.3}  & 
    \multicolumn{2}{c}{63.4$\pm$14.4}   
\\
    \multicolumn{1}{c}{UNet}       & 
    \multicolumn{1}{c}{EfficientNet-b5} & 
    \multicolumn{1}{c|}{31.2M}  &  
    \multicolumn{2}{c}{\textbf{91.9$\pm$0.4}}  & 
    \multicolumn{2}{c}{\textbf{92.4$\pm$0.8}}  & 
    \multicolumn{2}{c}{\underline{82.6$\pm$0.2}}  & 
    \multicolumn{2}{c}{54.2$\pm$3.3}  & 
    \multicolumn{2}{c}{81.0$\pm$2.3}  & 
    \multicolumn{2}{c}{\underline{90.6$\pm$0.4}}  & 
    \multicolumn{2}{c}{82.3$\pm$10.0}   
\\
    \multicolumn{1}{c}{Segformer}       & 
    \multicolumn{1}{c}{mit-b1} & 
    \multicolumn{1}{c|}{13.7M}  &  
    \multicolumn{2}{c}{89.7$\pm$0.3}  & 
    \multicolumn{2}{c}{91.7$\pm$0.5}  & 
    \multicolumn{2}{c}{82.3$\pm$1.0}  & 
    \multicolumn{2}{c}{52.3$\pm$1.1}  & 
    \multicolumn{2}{c}{80.2$\pm$1.0}  & 
    \multicolumn{2}{c}{90.0$\pm$0.3}  & 
    \multicolumn{2}{c}{\underline{75.4$\pm$12.2}}   
\\
    \multicolumn{1}{c}{Segformer}       & 
    \multicolumn{1}{c}{mit-b3} & 
    \multicolumn{1}{c|}{44.6M}  &  
    \multicolumn{2}{c}{\underline{91.8$\pm$0.4}}  & 
    \multicolumn{2}{c}{\underline{92.4$\pm$0.8}}  & 
    \multicolumn{2}{c}{\underline{83.0$\pm$0.6}}  & 
    \multicolumn{2}{c}{\underline{56.6$\pm$2.8}}  & 
    \multicolumn{2}{c}{\underline{82.3$\pm$1.5}}  & 
    \multicolumn{2}{c}{\underline{91.1$\pm$0.3}}  & 
    \multicolumn{2}{c}{\underline{79.6$\pm$19.9}}   
\\
    \multicolumn{1}{c}{DPT}       & 
    \multicolumn{1}{c}{ViT-Small} & 
    \multicolumn{1}{c|}{52.2}  &  
    \multicolumn{2}{c}{90.8$\pm$0.2}  & 
    \multicolumn{2}{c}{89.8$\pm$0.6}  & 
    \multicolumn{2}{c}{81.6$\pm$0.4}  & 
    \multicolumn{2}{c}{52.4$\pm$1.0}  & 
    \multicolumn{2}{c}{80.6$\pm$1.4}  & 
    \multicolumn{2}{c}{89.9$\pm$3.5}  & 
    \multicolumn{2}{c}{72.6$\pm$11.9}   
\\
    \multicolumn{1}{c}{SegDINO$^\dagger$}       & 
    \multicolumn{1}{c}{ViT-Small} & 
    \multicolumn{1}{c|}{9.5M}  &  
    \multicolumn{2}{c}{85.3$\pm$0.7}  & 
    \multicolumn{2}{c}{86.0$\pm$0.6}  & 
    \multicolumn{2}{c}{74.6$\pm$1.1}  & 
    \multicolumn{2}{c}{48.2$\pm$1.1}  & 
    \multicolumn{2}{c}{76.0$\pm$1.3}  & 
    \multicolumn{2}{c}{84.4$\pm$0.9}  & 
    \multicolumn{2}{c}{37.2$\pm$9.7}   
\\ \hline
    \multicolumn{1}{c}{Ours$^\dagger$}       & 
    \multicolumn{1}{c}{ViT-Small} & 
    \multicolumn{1}{c|}{11.3M}  &  
    \multicolumn{2}{c}{90.5$\pm$0.1}  & 
    \multicolumn{2}{c}{90.6$\pm$0.1}  & 
    \multicolumn{2}{c}{81.4$\pm$0.3}  & 
    \multicolumn{2}{c}{\underline{56.5$\pm$1.5}}  & 
    \multicolumn{2}{c}{\underline{84.4$\pm$1.2}}  & 
    \multicolumn{2}{c}{89.7$\pm$0.2}  & 
    \multicolumn{2}{c}{60.6$\pm$20.7}   
\\
    \multicolumn{1}{c}{Ours}       & 
    \multicolumn{1}{c}{ViT-Small} & 
    \multicolumn{1}{c|}{32.9M}  &  
    \multicolumn{2}{c}{\underline{91.4$\pm$1.8}}  & 
    \multicolumn{2}{c}{91.7$\pm$0.9}  & 
    \multicolumn{2}{c}{\textbf{85.4$\pm$1.0}}  & 
    \multicolumn{2}{c}{\textbf{56.7$\pm$2.2}}  & 
    \multicolumn{2}{c}{\textbf{86.7$\pm$1.0}}  & 
    \multicolumn{2}{c}{\textbf{91.5$\pm$0.6}}  & 
    \multicolumn{2}{c}{\textbf{95.5$\pm$1.3}}   
\\
\hline
\end{tabularx}%
}
\end{table}

\begin{figure}[t]
    \centering
    \includegraphics[width=1.0\linewidth]{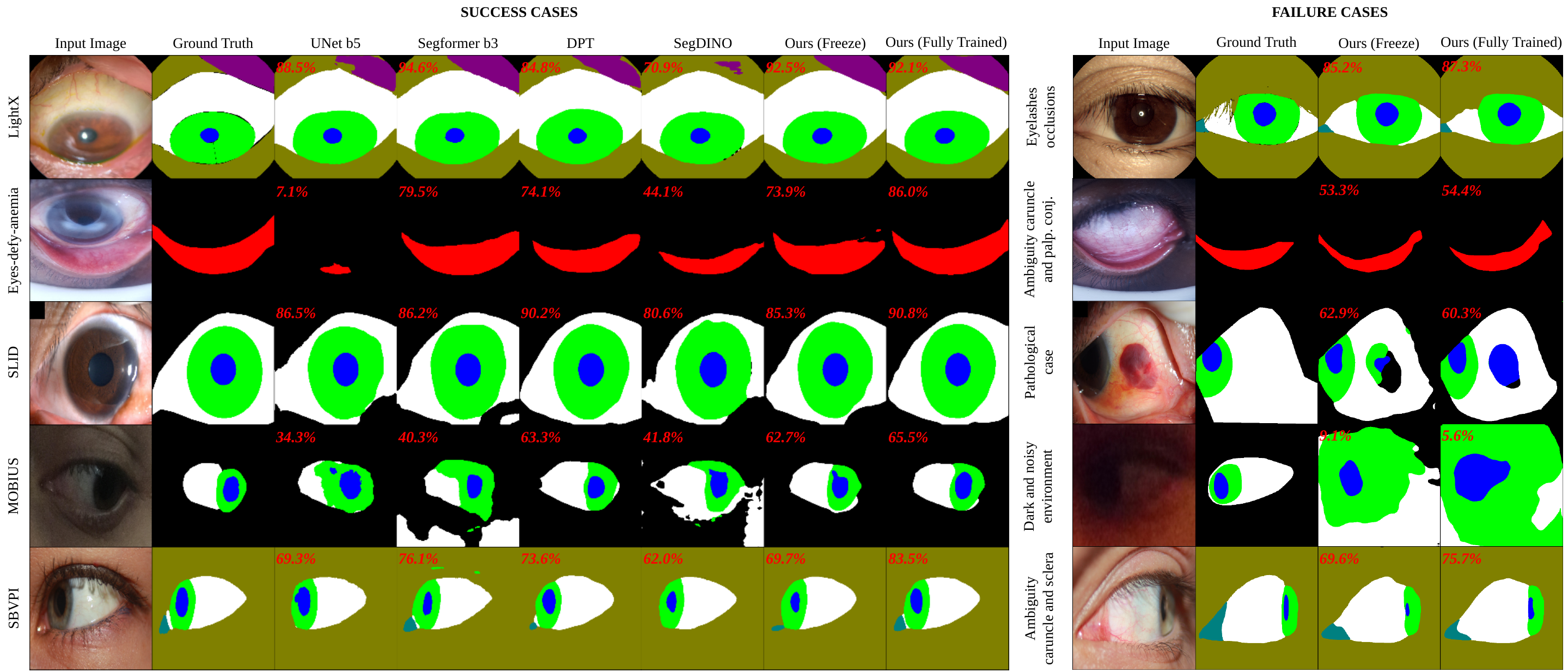}
    \caption{Qualitative segmentation results for the proposed method and competing baselines on representative success and failure cases. Corresponding mIoU scores are shown for each prediction.}
    \label{fig:viz}
\end{figure}

 \subsubsection{Data.} 
 We trained and evaluated our approach on  our private dataset LightX, consisting of 333 high-resolution AES images collected from 189 patients, after ethics review board approval (CER-2223-75-D). The dataset combines four sources, including EyePACS AES \cite{eyepacs} and three clinical collections acquired with smartphone-based slit-lamp systems across eight ophthalmic examination protocols. All images were manually annotated for seven classes: pupil, iris, sclera, palpebral conjunctiva, caruncle, periocular region, and acquisition artifacts. Each subset was split at the patient level into 60\% training, 20\% validation, and 20\% test sets before merging. To assess robustness to domain shifts, we additionally evaluate on four unseen public datasets: Eyes-Defy-Anemia \cite{eda} (218 images, including palpebral conjunctiva annotations only), SLID \cite{slid} (1,723 images, including pupil, iris and sclera annotations), MOBIUS \cite{mobius} (3,559 images, including pupil, iris and sclera annotations), and SBVPI \cite{sbvpi} (1,840 images, including all structures scattered across samples except artifacts). These datasets span both clinical and biometric settings and exhibit substantial variability in acquisition conditions and annotation protocols.

\subsubsection{Baselines.}
We compare the proposed approach against several representative segmentation architectures. As a convolutional baseline, we consider the classical UNet \cite{unet} equipped with both EfficientNet-B0 and EfficientNet-B5 encoders \cite{efficientnet}. As a lightweight transformer-based baseline, we include SegFormer \cite{xie2021segformer} with MiT-B1 and MiT-B3 backbones. We further compare against DPT \cite{Ranftl2021dpt} and SegDINO \cite{yang2025segdino},  which constitute the closest architectures to the proposed method. For a fair comparison, both DPT and SegDINO are implemented using the same DINOv3 pre-trained ViT-Small backbone as our approach, ensuring that performance differences primarily reflect the decoding strategy rather than the underlying feature extractor.

\subsubsection{Implementation details}
All models were trained using a combination of cross-entropy loss and generalized Dice loss \cite{gdice}, the latter mitigating class imbalance by weighting each class by its inverse squared frequency. Training was performed on the LightX dataset and evaluated on official test sets when available. Inputs were resized to $512\times512$ and augmented with random color jittering, Gaussian noise, horizontal flips and rotations. Models were trained for 200 epochs with a batch size of 8 using AdamW \cite{adamw}. The learning rate was selected per architecture in the range $[10^{-5},10^{-3}]$, with cosine annealing to $10^{-6}$. The best checkpoint was selected based on validation performance. For DPT, SegDINO, and our method, features were extracted from the 3rd, 6th, 9th, and 12th transformer layers. All experiments were run on a single NVIDIA RTX 4090 GPU and repeated over 5 random seeds, with results reported as averaged scores. Performance was evaluated using mean IoU (mIoU) and Dice similarity coefficient (DSC). 

\subsubsection{Comparison study.}

Global performance results are reported in Table \ref{tab:global-perf}. First lets consider results on in-domain data (LightX). UNet and SegFormer remain strong baselines, both reaching approximately 82.14\% mIoU. In contrast, DINOv3-based approaches struggle to fully exploit pre-trained representations. SegDINO achieves only 70.24\% mIoU, while DPT improves performance at the cost of a substantially larger decoder, yet remains below UNet and SegFormer. Our proposed architecture bridges this gap. With a frozen encoder, it significantly outperforms SegDINO and reaches DPT performance with nearly five times fewer parameters, highlighting the importance of decoder design when adapting foundation model representations to dense prediction tasks. When fully fine-tuned, as in DPT, our method achieves the best overall performance, reaching 85.55\% mIoU while maintaining a model size comparable to the UNet baseline. Per-class results reported in Table \ref{tab:per-class} further reveal the impact of annotation imbalance. Under-represented structures, including the caruncle, palpebral conjunctiva, and artifacts, exhibit both lower mIoU and higher standard deviations than frequently observed classes. The \textit{Artifacts} class is particularly challenging due to its scarcity and large visual variability.

Out-of-domain results further support these findings. Overall, ViT-based methods exhibit greater robustness to domain shifts than convolutional approaches, with our model consistently achieving the best cross-dataset performance. Nevertheless, a substantial domain gap remains, particularly on MOBIUS, whose acquisition conditions differ markedly from the training data.

Figure \ref{fig:viz} shows representative success and failure cases. While our approach produces robust segmentations across diverse domains, failures still occur under challenging acquisition conditions, including variations in imaging protocols and illumination (MOBIUS), the presence of large pathological structures (SLID), ambiguities between over- and under-represented classes (Eyes-defy-anemia and SBVPI) and fine-grained occlusions (LightX).

\subsubsection{Ablation study}
Results of the ablation study are reported in Table \ref{tab:ab-arch}. We assess the contribution of the step-attention feature refinement (SAFR) module, the convolutional decoder (ConvUp), and the use of RoPE and the class token (CLS) during feature refinement. Our final approach achieves the best performance, demonstrating that the proposed combination of attention-based refinement and lightweight convolutional decoding is well suited to AES segmentation. Interestingly, removing the class token greatly improves performance, suggesting that its global image-level representation may be less relevant for dense pixel-wise prediction tasks such as semantic segmentation.

\begin{table}[t]
\centering
\caption{Ablation results on key architectural components with DINOv3 ViT-Small frozen weights and evaluated on our pirvate dataset. First line corresponds to our proposed approach.}
\label{tab:ab-arch}
\resizebox{0.65\columnwidth}{!}{%
\begin{tabularx}{0.75\textwidth}{XXXX|X}
\hline
    \multicolumn{1}{c}{CLS} & 
    \multicolumn{1}{c}{RoPE} & 
    \multicolumn{1}{c}{ConvUp} & 
    \multicolumn{1}{c|}{SAFR} & 
    \multicolumn{1}{c}{mIoU}      
\\ 
\hline
  &  
  \multicolumn{1}{c}{\checkmark }  &
  \multicolumn{1}{c}{\checkmark }  &    
  \multicolumn{1}{c|}{\checkmark }    &      
  \multicolumn{1}{c}{\textbf{79.09 $\pm$ 1.65}}    
\\
    \multicolumn{1}{c}{\checkmark }  &  
    \multicolumn{1}{c}{\checkmark }  &   
    \multicolumn{1}{c}{\checkmark }     &
    \multicolumn{1}{c|}{\checkmark }  &
    \multicolumn{1}{c}{76.90 $\pm$ 1.38}    
\\

    \multicolumn{1}{c}{\checkmark }  &  
      &   
    \multicolumn{1}{c}{\checkmark }     &
    \multicolumn{1}{c|}{\checkmark }  &
    \multicolumn{1}{c}{75.80 $\pm$ 2.60}    
\\
  \multicolumn{1}{c}{\checkmark }   &  
  \multicolumn{1}{c}{\checkmark }    &       
  &  
  \multicolumn{1}{c|}{\checkmark }  &
  \multicolumn{1}{c}{75.06 $\pm$ 0.42}   
\\ 
    \multicolumn{1}{c}{\checkmark }   &     
  \multicolumn{1}{c}{\checkmark }   &  
  \multicolumn{1}{c}{\checkmark }    &       
  &  
  \multicolumn{1}{c}{69.78 $\pm$ 1.29}   
\\
\hline
\end{tabularx}%
}
\end{table}

\section{Conclusion}

In this work, we proposed a DINOv3-based approach for AES segmentation in clinical settings. Our method consistently outperforms strong baselines on both in- and out-of-domain data when fully fine-tuned, while remaining parameter-efficient, highlighting the importance of decoder design. Nevertheless, domain shift remains a key limitation not fully addressed by architectural improvements alone. These results further emphasize the value of pre-training and the potential for foundation models tailored to AES imaging.

\begin{credits}
\subsubsection{\ackname} We acknowledge Dr. Matej Vitek of University of Ljubljana, Slovenia for allowing us to use the \textit{SBVPI} and \textit{MOBIUS} datasets in this work. We acknowledge the financial support of the Natural Sciences and Engineering Research Council of Canada (NSERC) and of the MEDTEQ+ industrial consortium for research and innovation in healthcare technologies [ALLRP 608097-25].

\subsubsection{\discintname}
The authors have no competing interests to declare that are
relevant to the content of this article.
\end{credits}

\bibliographystyle{splncs04}
\bibliography{bibliography}

@article{eyepacs,
  title = {EyePACS: An Adaptable Telemedicine System for Diabetic Retinopathy Screening},
  volume = {3},
  ISSN = {1932-2968},
  url = {http://dx.doi.org/10.1177/193229680900300315},
  DOI = {10.1177/193229680900300315},
  number = {3},
  journal = {Journal of Diabetes Science and Technology},
  publisher = {SAGE Publications},
  author = {Cuadros,  Jorge and Bresnick,  George},
  year = {2009},
  month = may,
  pages = {509–516}
}

@data{eda,
doi = {10.21227/t5s2-4j73},
url = {https://dx.doi.org/10.21227/t5s2-4j73},
author = {Giovanni Dimauro and Rosalia Maglietta and Thulasi Bai and Sivachandar Kasiviswanathan},
publisher = {IEEE Dataport},
title = {Eyes-defy-anemia},
year = {2022} }

@incollection{sbvpi,
	title={Deep Sclera Segmentation and Recognition},
	author={Rot, Peter and Vitek, Matej and Grm, Klemen and Emer\v{s}i\v{c}, \v{Z}iga and Peer, Peter and \v{S}truc, Vitomir},
	booktitle={Handbook of Vascular Biometrics (HVB)},
	editor={Uhl, Andreas and Busch, Christoph and Marcel, S\'{e}bastien and Veldhuis, Raymond N. J.},
	pages={395--432},
	year={2020},
	publisher={Springer},
	doi="10.1007/978-3-030-27731-4_13"
}

@inproceedings{mobius,
	title={{SSBC} 2020: Sclera Segmentation Benchmarking Competition in the Mobile Environment},
	author={Vitek, Matej and Das, Abhijit and Pourcenoux, Yann and Missler, Alexandre and Paumier, Calvin and Das, Sumanta and De Ghosh, Ishita and Lucio, Diego R. and Zanlorensi Jr., Luiz A. and Menotti, David and Boutros, Fadi and Damer, Naser and Grebe, Jonas Henry and Kuijper, Arjan and Hu, Junxing and He, Yong and Wang, Caiyong and Liu, Hongda and Wang, Yunlong and Sun, Zhenan and Osorio-Roig, Daile and Rathgeb, Christian and Busch, Christoph and Tapia Farias, Juan and Valenzuela, Andres and Zampoukis, Georgios and Tsochatzidis, Lazaros and Pratikakis, Ioannis and Nathan, Sabari and Suganya, R and Mehta, Vineet and Dhall, Abhinav and Raja, Kiran and Gupta, Gourav and Khiarak, Jalil Nourmohammadi and Akbari-Shahper, Mohsen and Jaryani, Farhang and Asgari-Chenaghlu, Meysam and Vyas, Ritesh and Dakshit, Sristi and Dakshit, Sagnik and Peer, Peter and Pal, Umapada and \v{S}truc, Vitomir},
	booktitle={IEEE International Joint Conference on Biometrics (IJCB)},
	pages={1--10},
	year={2020},
	month={10},
	doi="10.1109/IJCB48548.2020.9304881"
}

@article{slid,
  title = {SLID: a slit-lamp image dataset for deep learning-based anterior eye anatomical segmentation and multi-lesion detection},
  volume = {7},
  ISSN = {2673-253X},
  url = {http://dx.doi.org/10.3389/fdgth.2025.1716501},
  DOI = {10.3389/fdgth.2025.1716501},
  journal = {Frontiers in Digital Health},
  publisher = {Frontiers Media SA},
  author = {Xu,  Mingyu and Sun,  Yiming and Cheng,  Huimin and Zhou,  Yifan and Maimaiti,  Nuliqiman and Chen,  Pengjie and Miao,  Qi and Xu,  Peifang and Ye,  Juan},
  year = {2026},
  month = Jan 
}

@misc{casia_irisv4,
url = {https://hycasia.github.io/dataset/casia-irisv4/},
author = {{Chinese Academy of Sciences' Institute of Automation (CASIA)}},
title = {CASIA Iris Image Database version 4.0},
year={2018}}

@article{CAMPOREALE2025111026,
title = {Highly reliable personalized noninvasive hemoglobin estimation by using Vision Transformers and dual fine-tuning},
journal = {Computers in Biology and Medicine},
volume = {197},
pages = {111026},
year = {2025},
issn = {0010-4825},
doi = {https://doi.org/10.1016/j.compbiomed.2025.111026},
url = {https://www.sciencedirect.com/science/article/pii/S0010482525013782},
author = {Mauro Camporeale and Felice Clemente and Giovanni Dimauro and Nunzia Lomonte and Rosalia Maglietta and Crescenza Pasciolla and Davide Sacco and Gian Maria Zaccaria},
}

@article{Nahass2025,
  title = {Open-Source Periorbital Segmentation Dataset for Ophthalmic Applications},
  volume = {5},
  ISSN = {2666-9145},
  url = {http://dx.doi.org/10.1016/j.xops.2025.100757},
  DOI = {10.1016/j.xops.2025.100757},
  number = {4},
  journal = {Ophthalmology Science},
  publisher = {Elsevier BV},
  author = {Nahass,  George R. and Koehler,  Emma and Tomaras,  Nicholas and Lopez,  Danny and Cheung,  Madison and Palacios,  Alexander and Peterson,  Jeffrey C. and Hubschman,  Sasha and Green,  Kelsey and Purnell,  Chad A. and Setabutr,  Pete and Tran,  Ann Q. and Yi,  Darvin},
  year = {2025},
  month = July,
  pages = {100757}
}

@article{Wong2025,
  title = {Toward automated assessment of conjunctival hyperemia: A semisupervised artificial intelligence approach},
  volume = {1551},
  ISSN = {1749-6632},
  url = {http://dx.doi.org/10.1111/nyas.70009},
  DOI = {10.1111/nyas.70009},
  number = {1},
  journal = {Annals of the New York Academy of Sciences},
  publisher = {Wiley},
  author = {Wong,  Damon and Ng,  Yvonne and Eppenberger,  Leila Sara and Cherecheanu,  Alina Popa and Anghelache,  Anca and Toma,  Eduard and Coroleuca,  Ruxandra and Garcia‐Feijoo,  Julian and Garh\"{o}fer,  Gerhard and Schmetterer,  Leopold},
  year = {2025},
  month = Aug,
  pages = {201–209}
}

@article{Visionome,
  title = {Dense anatomical annotation of slit-lamp images improves the performance of deep learning for the diagnosis of ophthalmic disorders},
  volume = {4},
  ISSN = {2157-846X},
  url = {http://dx.doi.org/10.1038/s41551-020-0577-y},
  DOI = {10.1038/s41551-020-0577-y},
  number = {8},
  journal = {Nature Biomedical Engineering},
  publisher = {Springer Science and Business Media LLC},
  author = {Li,  Wangting and Yang,  Yahan and Zhang,  Kai and Long,  Erping and He,  Lin and Zhang,  Lei and Zhu,  Yi and Chen,  Chuan and Liu,  Zhenzhen and Wu,  Xiaohang and Yun,  Dongyuan and Lv,  Jian and Liu,  Yizhi and Liu,  Xiyang and Lin,  Haotian},
  year = {2020},
  month = jun,
  pages = {767–777}
}

@misc{ssbc2025,
      title={Privacy-enhancing Sclera Segmentation Benchmarking Competition: SSBC 2025}, 
      author={Matej Vitek and Darian Tomašević and Abhijit Das and Sabari Nathan and Gökhan Özbulak and Gözde Ayşe Tataroğlu Özbulak and Jean-Paul Calbimonte and André Anjos and Hariohm Hemant Bhatt and Dhruv Dhirendra Premani and Jay Chaudhari and Caiyong Wang and Jian Jiang and Chi Zhang and Qi Zhang and Iyyakutti Iyappan Ganapathi and Syed Sadaf Ali and Divya Velayudan and Maregu Assefa and Naoufel Werghi and Zachary A. Daniels and Leeon John and Ritesh Vyas and Jalil Nourmohammadi Khiarak and Taher Akbari Saeed and Mahsa Nasehi and Ali Kianfar and Mobina Pashazadeh Panahi and Geetanjali Sharma and Pushp Raj Panth and Raghavendra Ramachandra and Aditya Nigam and Umapada Pal and Peter Peer and Vitomir Štruc},
      year={2025},
      eprint={2508.10737},
      archivePrefix={arXiv},
      primaryClass={cs.CV},
      url={https://arxiv.org/abs/2508.10737}, 
}

@article{yang2025segdino,
  title={SegDINO: An Efficient Design for Medical and Natural Image Segmentation with DINO-V3},
  author={Yang, Sicheng and Wang, Hongqiu and Xing, Zhaohu and Chen, Sixiang and Zhu, Lei},
  journal={arXiv preprint arXiv:2509.00833},
  year={2025},
  url={https://arxiv.org/abs/2509.00833}
}

@article{Ranftl2021dpt,
	author    = {Ren\'{e} Ranftl and Alexey Bochkovskiy and Vladlen Koltun},
	title     = {Vision Transformers for Dense Prediction},
	journal   = {ArXiv preprint},
	year      = {2021},
}

@misc{simeoni2025dinov3,
  title={{DINOv3}},
  author={Sim{\'e}oni, Oriane and Vo, Huy V. and Seitzer, Maximilian and Baldassarre, Federico and Oquab, Maxime and Jose, Cijo and Khalidov, Vasil and Szafraniec, Marc and Yi, Seungeun and Ramamonjisoa, Micha{\"e}l and Massa, Francisco and Haziza, Daniel and Wehrstedt, Luca and Wang, Jianyuan and Darcet, Timoth{\'e}e and Moutakanni, Th{\'e}o and Sentana, Leonel and Roberts, Claire and Vedaldi, Andrea and Tolan, Jamie and Brandt, John and Couprie, Camille and Mairal, Julien and J{\'e}gou, Herv{\'e} and Labatut, Patrick and Bojanowski, Piotr},
  year={2025},
  eprint={2508.10104},
  archivePrefix={arXiv},
  primaryClass={cs.CV},
  url={https://arxiv.org/abs/2508.10104},
}

@inproceedings{caron2021dino,
  title={Emerging Properties in Self-Supervised Vision Transformers},
  author={Caron, Mathilde and Touvron, Hugo and Misra, Ishan and J\'egou, Herv\'e  and Mairal, Julien and Bojanowski, Piotr and Joulin, Armand},
  booktitle={Proceedings of the International Conference on Computer Vision (ICCV)},
  year={2021}
}

@InProceedings{unet,
author="Ronneberger, Olaf
and Fischer, Philipp
and Brox, Thomas",
editor="Navab, Nassir
and Hornegger, Joachim
and Wells, William M.
and Frangi, Alejandro F.",
title="U-Net: Convolutional Networks for Biomedical Image Segmentation",
booktitle="Medical Image Computing and Computer-Assisted Intervention -- MICCAI 2015",
year="2015",
publisher="Springer International Publishing",
address="Cham",
pages="234--241",
}

@misc{rope,
      title={RoFormer: Enhanced Transformer with Rotary Position Embedding}, 
      author={Jianlin Su and Yu Lu and Shengfeng Pan and Ahmed Murtadha and Bo Wen and Yunfeng Liu},
      year={2023},
      eprint={2104.09864},
      archivePrefix={arXiv},
      primaryClass={cs.CL},
      url={https://arxiv.org/abs/2104.09864}, 
}

@InProceedings{gdice,
author="Sudre, Carole H.
and Li, Wenqi
and Vercauteren, Tom
and Ourselin, Sebastien
and Jorge Cardoso, M.",
editor="Cardoso, M. Jorge
and Arbel, Tal
and Carneiro, Gustavo
and Syeda-Mahmood, Tanveer
and Tavares, Jo{\~a}o Manuel R.S.
and Moradi, Mehdi
and Bradley, Andrew
and Greenspan, Hayit
and Papa, Jo{\~a}o Paulo
and Madabhushi, Anant
and Nascimento, Jacinto C.
and Cardoso, Jaime S.
and Belagiannis, Vasileios
and Lu, Zhi",
title="Generalised Dice Overlap as a Deep Learning Loss Function for Highly Unbalanced Segmentations",
booktitle="Deep Learning in Medical Image Analysis and Multimodal Learning for Clinical Decision Support ",
year="2017",
publisher="Springer International Publishing",
address="Cham",
pages="240--248",
isbn="978-3-319-67558-9"
}

@misc{efficientnet,
      title={EfficientNet: Rethinking Model Scaling for Convolutional Neural Networks}, 
      author={Mingxing Tan and Quoc V. Le},
      year={2020},
      eprint={1905.11946},
      archivePrefix={arXiv},
      primaryClass={cs.LG},
      url={https://arxiv.org/abs/1905.11946}, 
}

@inproceedings{xie2021segformer,
  title={SegFormer: Simple and Efficient Design for Semantic Segmentation with Transformers},
  author={Xie, Enze and Wang, Wenhai and Yu, Zhiding and Anandkumar, Anima and Alvarez, Jose M and Luo, Ping},
  booktitle={Neural Information Processing Systems (NeurIPS)},
  year={2021}
}

@misc{adamw,
      title={Decoupled Weight Decay Regularization}, 
      author={Ilya Loshchilov and Frank Hutter},
      year={2019},
      eprint={1711.05101},
      archivePrefix={arXiv},
      primaryClass={cs.LG},
      url={https://arxiv.org/abs/1711.05101}, 
}

@inproceedings{chaudhary2019ritnet,
  title={RITnet: real-time semantic segmentation of the eye for gaze tracking},
  author={Chaudhary, Aayush K and Kothari, Rakshit and Acharya, Manoj and Dangi, Shusil and Nair, Nitinraj and Bailey, Reynold and Kanan, Christopher and Diaz, Gabriel and Pelz, Jeff B},
  booktitle={2019 IEEE/CVF International Conference on Computer Vision Workshop (ICCVW)},
  pages={3698--3702},
  year={2019},
  organization={IEEE}
}
\end{document}